\documentclass[runningheads]{llncs}
\usepackage[T1]{fontenc}
\usepackage{graphicx}
\usepackage{siunitx}
\usepackage{algpseudocode}
\usepackage{algorithm}
\usepackage{graphicx}
\usepackage{multirow}
\usepackage{adjustbox}
\usepackage{booktabs}
\usepackage{makecell} 
\usepackage{float}

\begin{document}
\title{An Open-source Capping Machine Suitable for Confined
Spaces}
%
%
\author{Francisco Munguia-Galeano\inst{1}\orcidID{0000-0001-8397-3083} \and
Louis Longley\inst{1}\orcidID{0000-0002-9178-9603} \and
Satheeshkumar Veeramani\inst{1}\orcidID{0000-0003-2538-002}
\and
Zhengxue Zhou\inst{1}\orcidID{0000-0001-9478-9361}
\and
Rob Clowes\inst{1}\orcidID{0009-0002-4686-2119}
\and
Hatem Fakhruldeen\inst{1}\orcidID{0009-0004-5043-2159}
\and
Andrew I. Cooper\inst{1}\orcidID{0000-0003-0201-1021}}
\authorrunning{F. Munguia-Galeano et al.}
%
\institute{Cooper Group, Department of Chemistry, University of Liverpool, United Kingdom\\
Corresponding Author: \email{aicooper@liverpool.ac.uk}}
\maketitle              
\begin{abstract}
In the context of self-driving laboratories (SDLs), ensuring automated and error-free capping is crucial, as it is a ubiquitous step in sample preparation. Automated capping in SDLs can occur in both large and small workspaces (\textit{e.g.}, inside a fume hood). However, most commercial capping machines are designed primarily for large spaces and are often too bulky for confined environments. Moreover, many commercial products are closed-source, which can make their integration into fully autonomous workflows difficult. This paper introduces an open-source capping machine suitable for compact spaces, which also integrates a vision system that recognises capping failure. The capping and uncapping processes are repeated 100 times each to validate the machine's design and performance. As a result, the capping machine reached a 100~\% success rate for capping and uncapping. Furthermore, the machine sealing capacities are evaluated by capping 12 vials filled with solvents of different vapour pressures: water, ethanol and acetone. The vials are then weighed every 3 hours for three days. The machine's performance is benchmarked against an industrial capping machine (a Chemspeed station) and manual capping. The vials capped with the prototype lost 0.54~\% of their content weight on average per day, while the ones capped with the Chemspeed and manually lost 0.0078~\% and 0.013~\%, respectively. The results show that the capping machine is a reasonable alternative to industrial and manual capping, especially when space and budget are limitations in SDLs.

\end{abstract}
\section{Introduction}

Self-driving laboratories (SDLs) are fundamental to tackling global challenges such as renewable energy, sustainability and healthcare and are pivotal for technological advancement in a variety of fields. In this context, the automation of experimentation and analysis, through the use of labware, specialised software and robotics, has been shown to be beneficial in both accelerating the discovery of new materials ~\cite{Li2015SmallMolecule,Salley2023Chemput} and their characterisation and formulation ~\cite{burger2020mobile,Wu2023Laser}. However, this automation process often requires expertise in both areas, leading to chemists facing challenges regarding the lack of specialised hardware that can fit the necessities and requirements of chemistry automation and roboticists lacking a sense of where and how automation can benefit chemistry.
Capping is ubiquitous and mandatory when preparing samples across various industry sectors. In chemistry~\cite{10093437}, capping and uncapping are necessary both during reaction workflows and for long-term safe storage and handling of vials. Nevertheless, the need for compact capping equipment becomes critical when workflows are conducted in a confined space, such as a fume hood, where the bulkiness of traditional industrial capping machines often makes them unsuitable.

\begin{figure}[b!]
    \centering
            \includegraphics[width=4.7in]{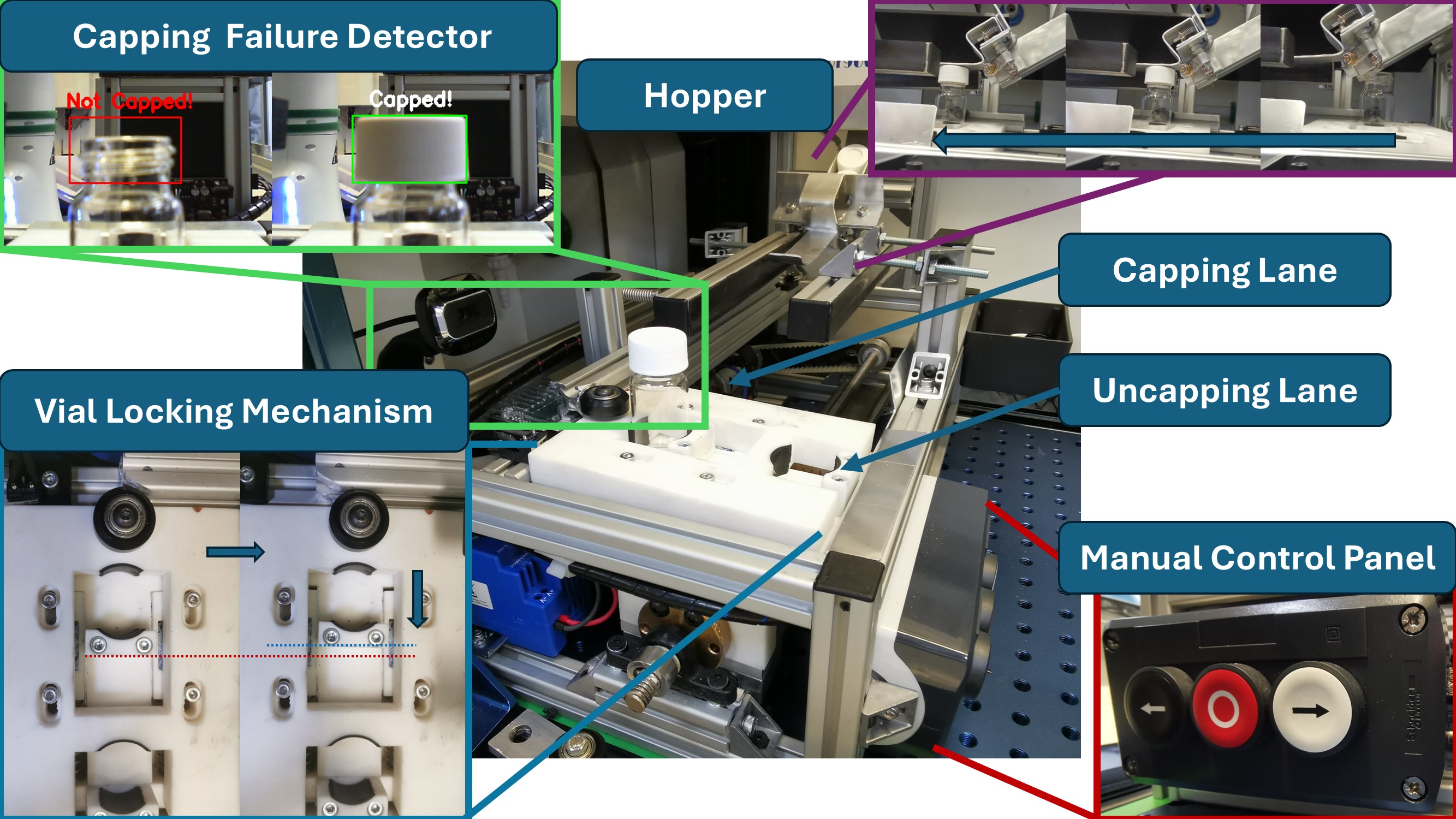}%
    \caption{Low-cost capping machine for confined spaces. Key components include a capping failure detector, a vial locking mechanism, a hopper, capping and uncapping lanes, and a manual control panel. The compact form factor allows deployment in constrained spaces such as fume hoods.}
    \label{fig:capper}
\end{figure}

To address these challenges, this paper introduces a capping machine (Fig.~\ref{fig:capper}) that incorporates vision-based capping failure detection and is specifically designed for use in constricted spaces.
It aims to be a low-cost alternative to industrial capping machines while being readily compatible with chemistry automation workflows. The capping machine's design and performance are validated through two experiments: capping success rate and sealing capacities. Both experiments are run in a fume hood, where a Panda robot manipulates the vials. The sealing capacities are benchmarked against two baselines: manual capping and an industrial capper. For the latest, a Chemspeed station (a common and well-known platform for chemistry automation) is used as the industrial capper.


\section{Related Work}
\label{sec:related_work}

In the literature, there exists a limited number of works discussing the design and control of capping machines. These works focus more on the control of such machines and their interfaces. For example, Avunoori Anudeep et al.~\cite{kumar2014automation} introduce a bottle-filling system that comprises a sequence of operations such as automatic clamping, unclamping, injecting the molten material, and filling and capping. In a similar work, Qiadong et al.~\cite{yao2023design} present a capping machine in which the force and torques necessary to cap bottles properly are validated through a finite element analysis while pressed by a pneumatic piston. It was concluded that the stress and pressure experienced by the cap threads are less than the yield strength of the material during the capping, producing low deformation and, at the same time, securing a robust capping.  

Other works focus more on the control design of the filling machines than on the design of the machines themselves. For example, Mahrez et al.~\cite{mahrez2022design} present a closed-loop control design for a bottle-filling capping system, in which the authors focus their aims on lower power consumption and operating costs. Several sensors detect the bottles' position and water level remaining in the tank while the capping process is achieved using a robotic arm. Another case in point is introduced by Zhang et al.~\cite{zhang2020control}, in which the authors motivate the advantages of improving the capping process of drug bottles because it shortens production time and saves labour. Like the previously cited approaches, the authors introduce a system that transports bottles to the capping module via a conveyor. The system then locks the bottle, and an actuator tightens the cap.  

Several studies have implemented uncapping. For instance,  Jaeger et al.~\cite{jaeger2021automated} introduce a system for uncapping multiple sample tubes. This solution is motivated by the fact that operators usually do the process manually, which is time-consuming and increases the risk of muscle fatigue and injuries. Another approach involves designing grippers that adapt according to the task; for example, Kumar et al.~\cite{kumar2016design} present the design of several grippers that are tailored to move vials and cap/uncap them. For decades, industry has thoroughly developed capping machines and utilised them primarily for bottle-filling applications. Conversely, the scientific literature presents a few designs focusing on low-cost builds or control and user interfaces. 

Industrial capping machines~\cite{ZoneSunCappers} combine liquid filling, cap feeding, and capping capabilities for various applications with customisable container sizes, processing up to 35 bottles per minute. Nevertheless, they cannot be directly integrated into SDLs due to their space requirements, and such a high productivity rate is unnecessary. Some manufacturers produce compact screw capping devices~\cite{AzentaIntellicap} for sample tubes; however, similar devices are not available for vials. Additionally, there is a gap concerning the lack of designs aimed at operating in automated chemistry workflows in confined spaces. This paper fills the gap by proposing an open-source capping machine suitable for confined spaces to help researchers and people involved in chemistry automation implement capping and uncapping reliably where setups involving reduced spaces are necessary. 

\begin{figure}[t]
    \centering
    \includegraphics[width=0.9\linewidth]{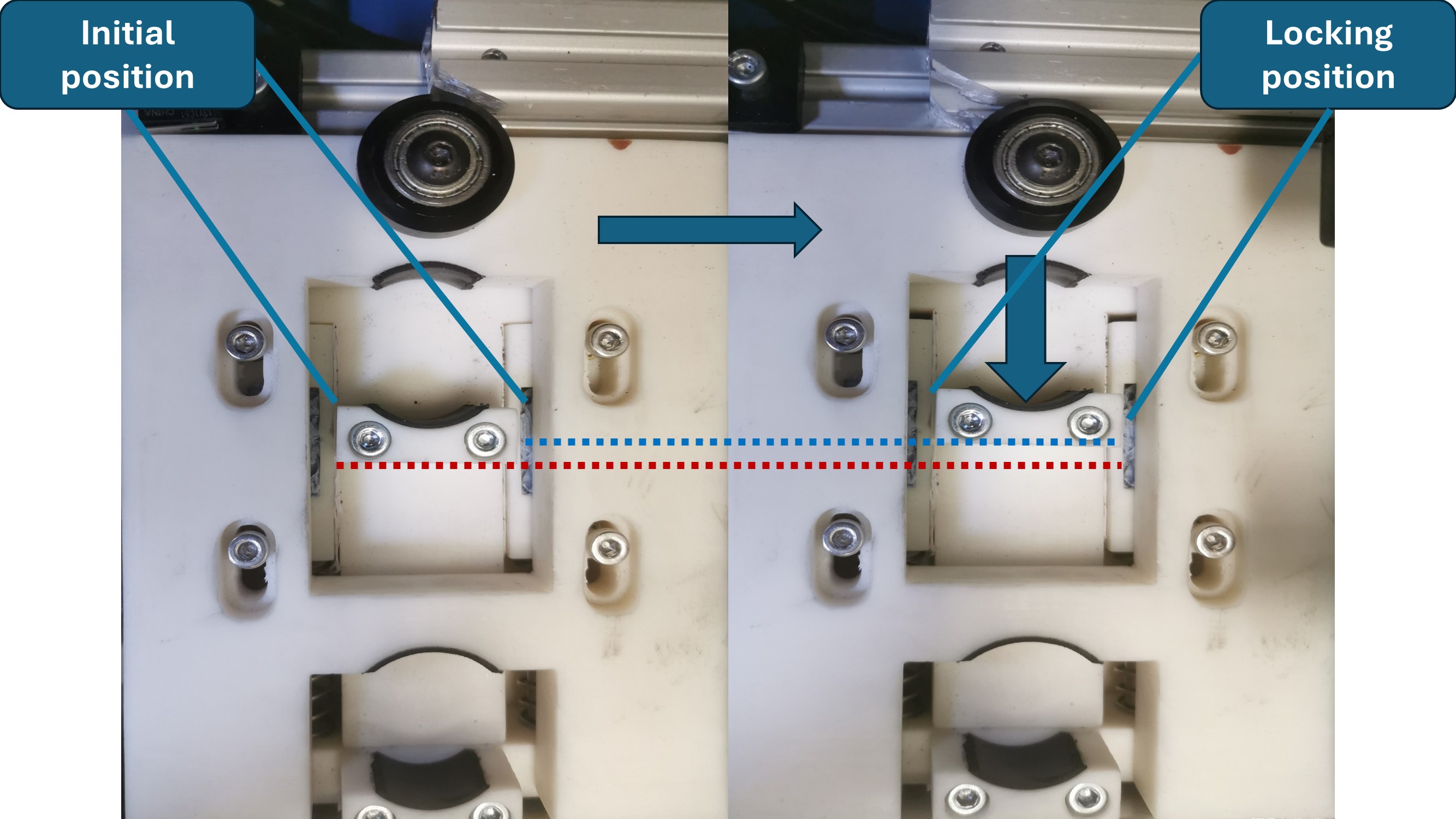}
    \caption{Vial holder and its locking mechanism. The image illustrates the engagement between the bearing and the cam when the vial holder moves to the left. This mechanism interaction causes the two fasteners to move towards one another, which locks the vial in place. The movement can be seen as the gap between the red and blue lines.}
    \label{fig:locking}
\end{figure}

\section{Capping Machine Design}
\label{sec:cmd}

This section presents the design of the capping machine, which encompasses three aspects: mechanics, electronics and firmware. These aspects are expected to meet the following system specifications: \textcircled{\tiny I} \textbf{Robustness} (\textit{i.e.}, the machine should reliably cap vials and detect when a failure occurs), \textcircled{\tiny II} \textbf{Easy to fabricate} (\textit{i.e.}, the machine's parts should be easy to replicate, and the electronic components should be easy to acquire), \textcircled{\tiny III} \textbf{Easy to use} (\textit{i.e.}, the machine should be easy to install, operate, and maintain), and \textcircled{\tiny IV} \textbf{Reduced size} (\textit{i.e.}, should fit in a fume hood, usually measuring approximately 760 mm to 1620 mm in width, around 580 mm to 670 mm in depth and typically 650 mm to 870 mm in height). To fulfil these requirements, the prototype design process is divided into three subsystems: mechanical, electrical/electronics and vision. 

\subsection{Mechanical subsystem}

\begin{figure}[b!]
    \centering
    \includegraphics[width=0.9\linewidth]{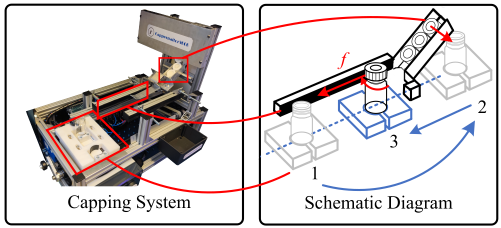}
    \caption{Capping system and its schematic diagram. First, the vial is placed when the vial holder is at position 1. Then, the vial holder moves to position 2. The capping process is achieved by utilising the friction between the cap and the rubber rail, which tightens the cap as it passes below the cap feeder on its way to the home position.}
    \label{fig:capping}
\end{figure}

When designing mechanisms, it is crucial to implement the concept of mechanical multiplexing, which consists of utilising a single input (from an actuator) and transforming it into multiple outputs. The output can include actions for locking, pushing, or tightening, and this behaviour is achieved by combining several mechanisms such as cams, gears, lead screws or linkages. Mechanical multiplexing is essential because it can cut costs and optimise efficiency by reducing the number of actuators and, thereby, the number of electric/electronic components to control a machine; the prototype of the capping machine presented here is designed based on this concept. It uses a motor to drive a lead screw via a timing belt. The lead screw converts the rotational motion of the motor into linear motion. A cam parallel to this linear motion is employed to engage the locking mechanism in the vial holder, allowing it to lock or unlock the vials (see Fig.~\ref{fig:locking}). The cap hopper and feeder are positioned above the vial holder. For the capping process, when a vial passes under the feeder, a cap is automatically placed on it. A profile with rubber positioned parallel to the vial cap spins the cap and tightens it, as shown in Fig.~\ref{fig:capping}. For the uncapping process, the capped vial must be placed in the uncapping lane, which operates on the same principle as the capping lane. However, the cap rotates in the uncapping lane in the opposite direction, and, as a consequence, the combination of movements loosens the cap.

The frame of the capping machine is constructed from 20x20 mm aluminium profiles, a common and well-known material widely used in automation setups. The dimensions of the prototype are 300\,mm in width, 500\,mm in depth, and 400\,mm in height---whereas an industrial capper from Chemspeed measures approximately 1000\,mm in width, 1000\,mm in depth, and 2200\,mm in height. The motor driving the lead screw is a brushed geared DC motor with 19.8 W of power, 12 V DC, 59 Ncm of torque, 84 rpm, and a 6 mm shaft. The lead screw has a shaft diameter of 10 mm and is fitted with a flanged round nut of the same diameter. The timing belt is a T5 type, with 68 teeth, 340 mm length, and 10 mm width. Two aluminium timing belt pulleys, each with 15 teeth, are used for a 10 mm wide belt and 5 mm pitch, maintaining a 1:1 ratio. The lead screw is supported by two pillow block bearings, one at each end. The vial holder is 3D-printed in resin and incorporates the lead screw nut, a linear bearing, two rods, a cam-engaging bearing, and two springs. The holder is mounted on an Igus linear guide carriage, model WW-10-40-10, which slides along an Igus W Series WS-10-40-600 linear guide rail, with a width of 40 mm.

\begin{figure*}[t!]
    \centering
    \includegraphics[width=1.0\linewidth]{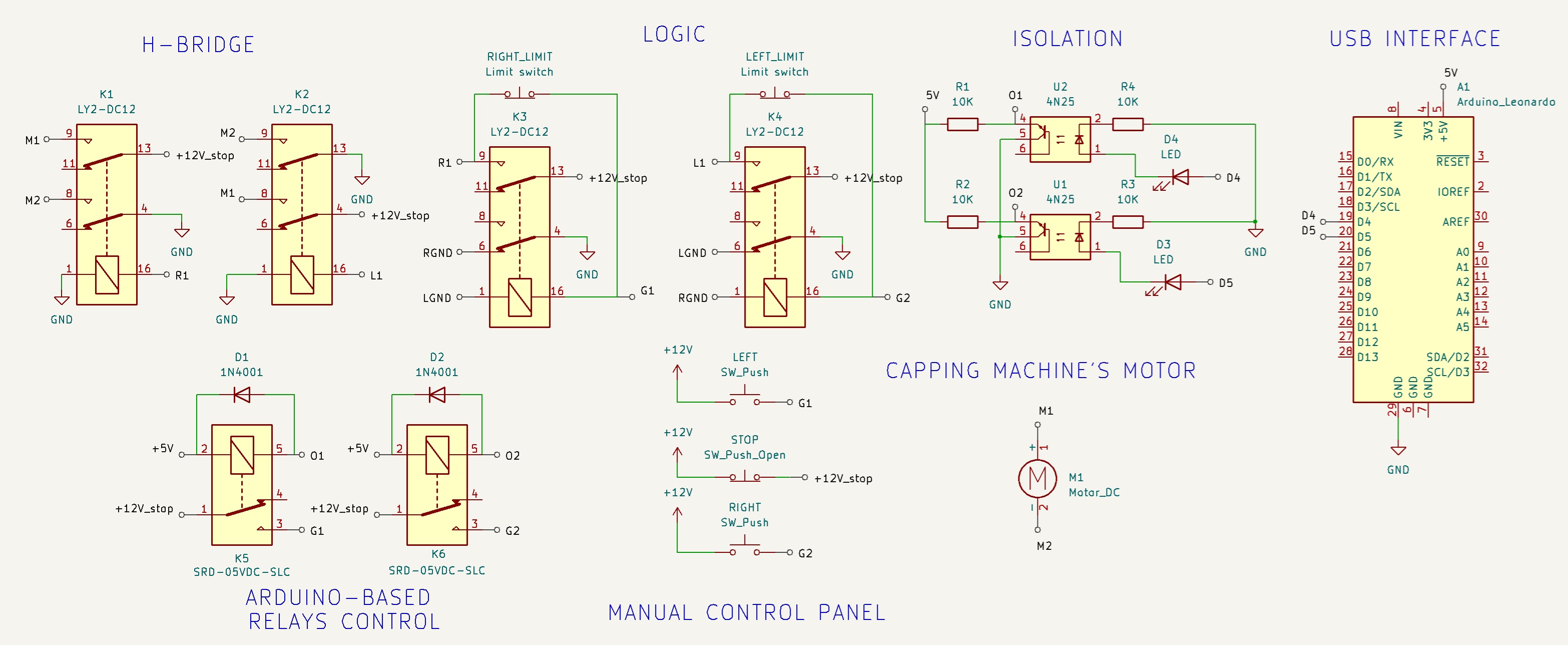}
    \caption{Arduino-based DC Motor Control Circuit Schematic of the Capping Machine. This circuit design ensures safe control of the DC motor by using a manual control panel and an Arduino controller with protection mechanisms. Relays are shown in their unpowered states.}
    \label{fig:schematic}
\end{figure*}

\begin{figure}[b!]
    \centering
            \includegraphics[width=4.4in]{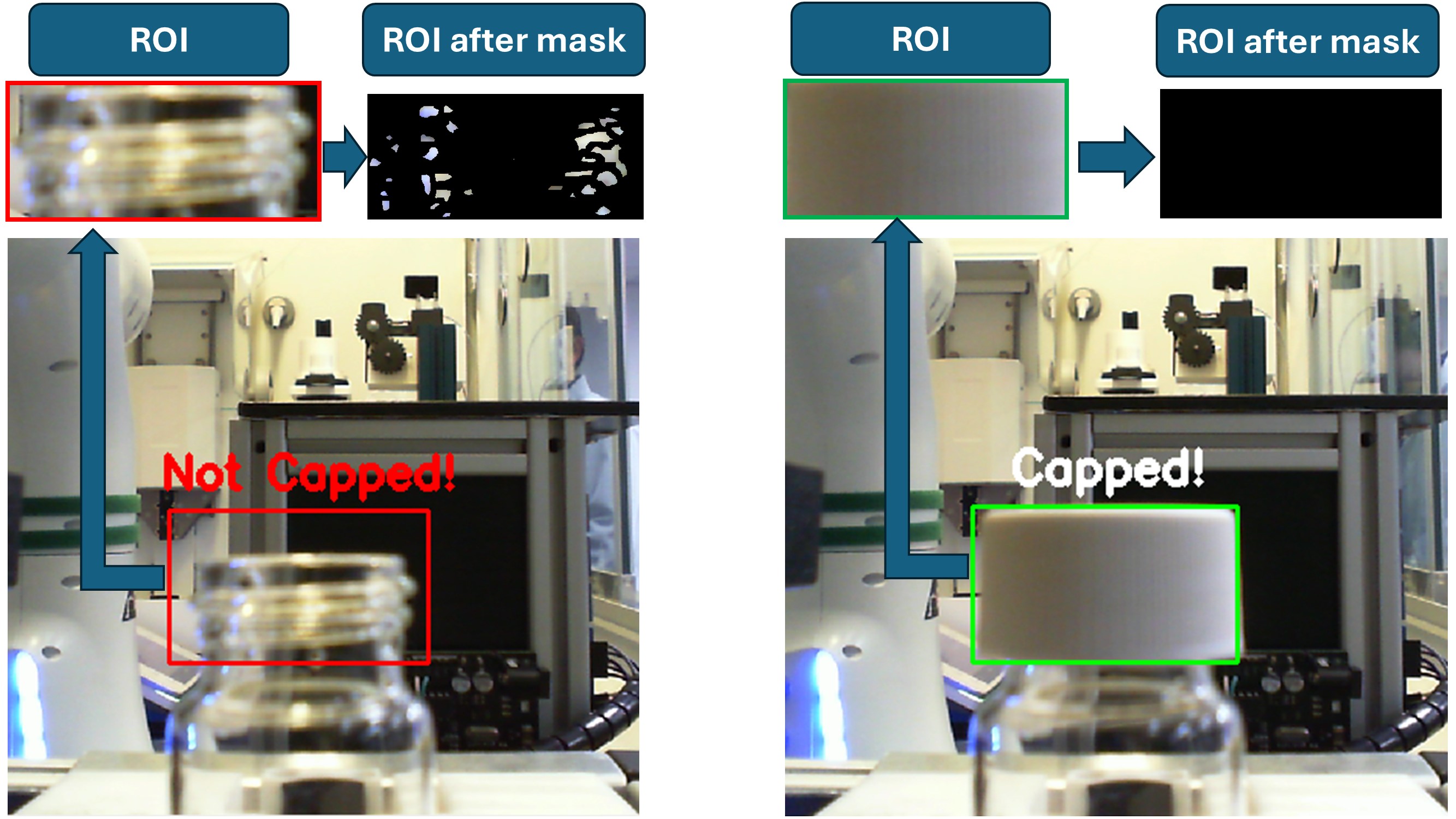}%
    \caption{Capping failure detector. On the left side of the image, the algorithm, after defining a region of interest (ROI), applies a mask and segments the light reflected through the vial's threads, indicating capping failure. On the right side of the image, after segmenting the ROI, the algorithm could not find any light reflected through the thread, meaning that the capping was successful.}
    \label{fig:visionmodule}
\end{figure}

\subsection{Electrical/Electronics subsystem}

Building on the mechanical system, the electrical/electronics subsystem controls the geared motor, provides a manual interface, and enables PC-based automation. An H-bridge formed by four LY2-DC12 relays allows bidirectional motor control. Two limit switches at the ends of the vial holder’s track toggle the relays based on direction. Manual control is provided via a spring-return push-button panel with three buttons: home, cap feeder, and emergency stop. PC control is handled by an Arduino Uno, which drives two SRD-05VDC-SL-C relays through an isolation stage with two 4N25 optocouplers and 10 k$\Omega$ resistors, protecting the board and providing LED feedback. Cable management uses conduit and spiral wrap. The system is powered by two DRL-30 switched supplies (85–264 V AC input, 12 V DC output, 2.1 A, 30 W), with all components mounted on a DIN rail. A detailed schematic is shown in Fig.~\ref{fig:schematic}.

\begin{algorithm}[t!]
\scriptsize
\caption{Capping Failure Detector Algorithm}
\begin{algorithmic}[1]
\State \textbf{Initialise} region of interest (ROI) coordinates
\While{Camera is streaming}
    \State Capture frame from camera
    \If{frame is not available}
        \State Display error and \textbf{exit loop}
    \EndIf
    \State Extract ROI from the frame
    \State Convert ROI to HSV
    
    \State Define colour range for filtering
    \State Apply colour mask to the ROI
    
    \State Convert the masked ROI to grayscale
    \State Apply thresholding to highlight relevant areas
    
    \State Find contours in the thresholded image
    \State Compute total area of detected contours
    
    \If{contour area exceeds threshold}
        \State Display "Not Capped" status and mark ROI in red
        \State Set capped status to \textbf{False}
    \Else
        \State Display "Capped" status and mark ROI in green
        \State Set capped status to \textbf{True}
    \EndIf
    
    \State Show processed frame in the display window
    \If{termination condition (\textit{e.g.}, key press) is met}
        \State \textbf{exit loop}
    \EndIf
\EndWhile
\end{algorithmic}
\end{algorithm}

\subsection{Vision subsystem}

The vision subsystem is a simple, reliable capping failure detector using a standard RGB camera. It leverages reflections from vial threads visible in the blue channel to segment the cap region. If the cap fully covers the threads, no blue regions are detected, indicating a successful seal . Conversely, visible blue areas above a defined threshold signal a failed capping attempt. This triggers an alert or stops the system for safety as depicted in Fig.~\ref{fig:visionmodule}. The step-by-step implementation of this approach is given by Algorithm 1. 

\section{Experimental Setup}
\label{sec:setup}

\begin{figure}[b!]
    \centering
            \includegraphics[width=4.0in]{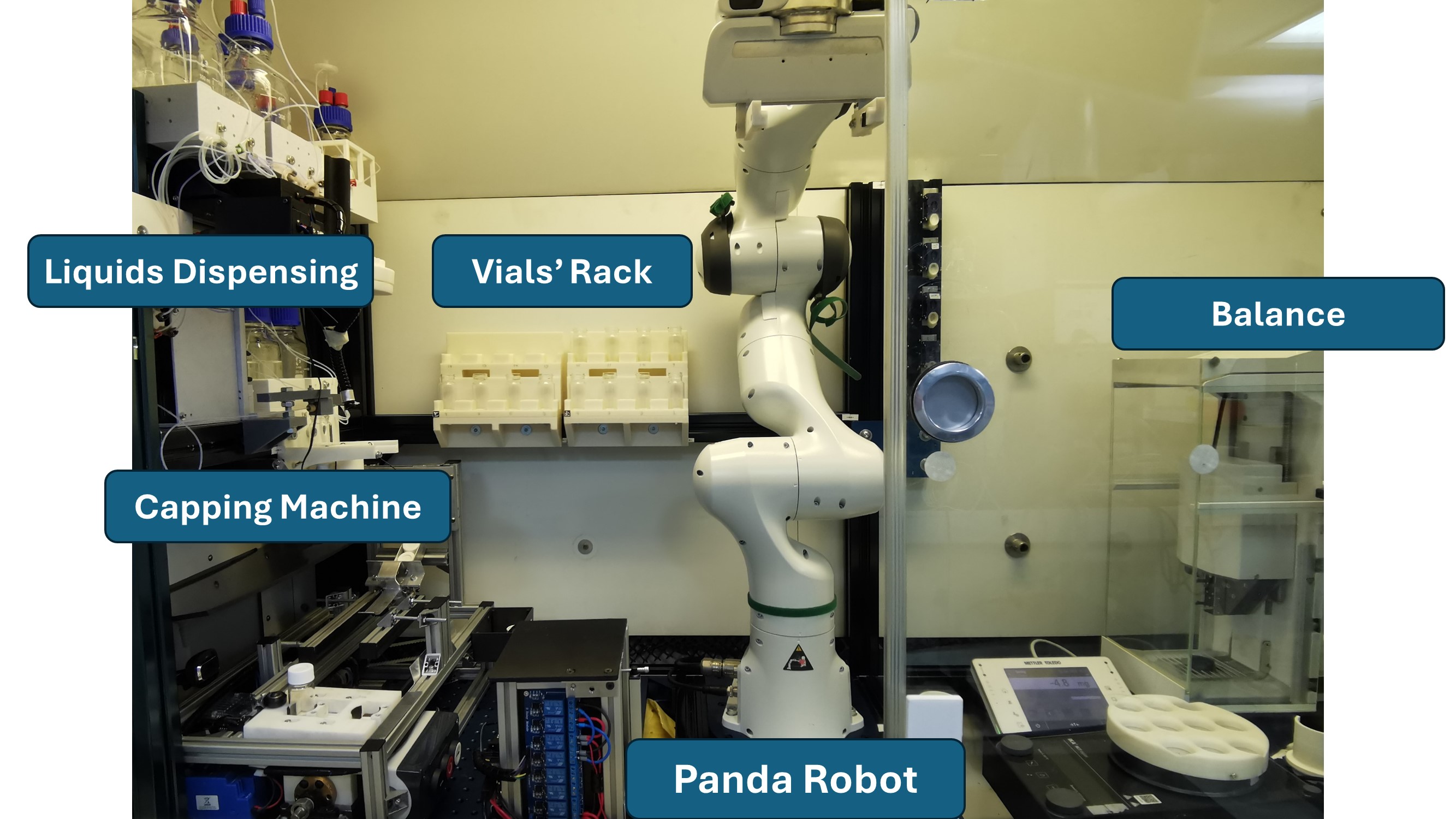}%
    \caption{Experimental setup.}
    \label{fig:expsetup}
\end{figure}

\begin{figure}[b!]
    \centering
    \includegraphics[width=4.4in]{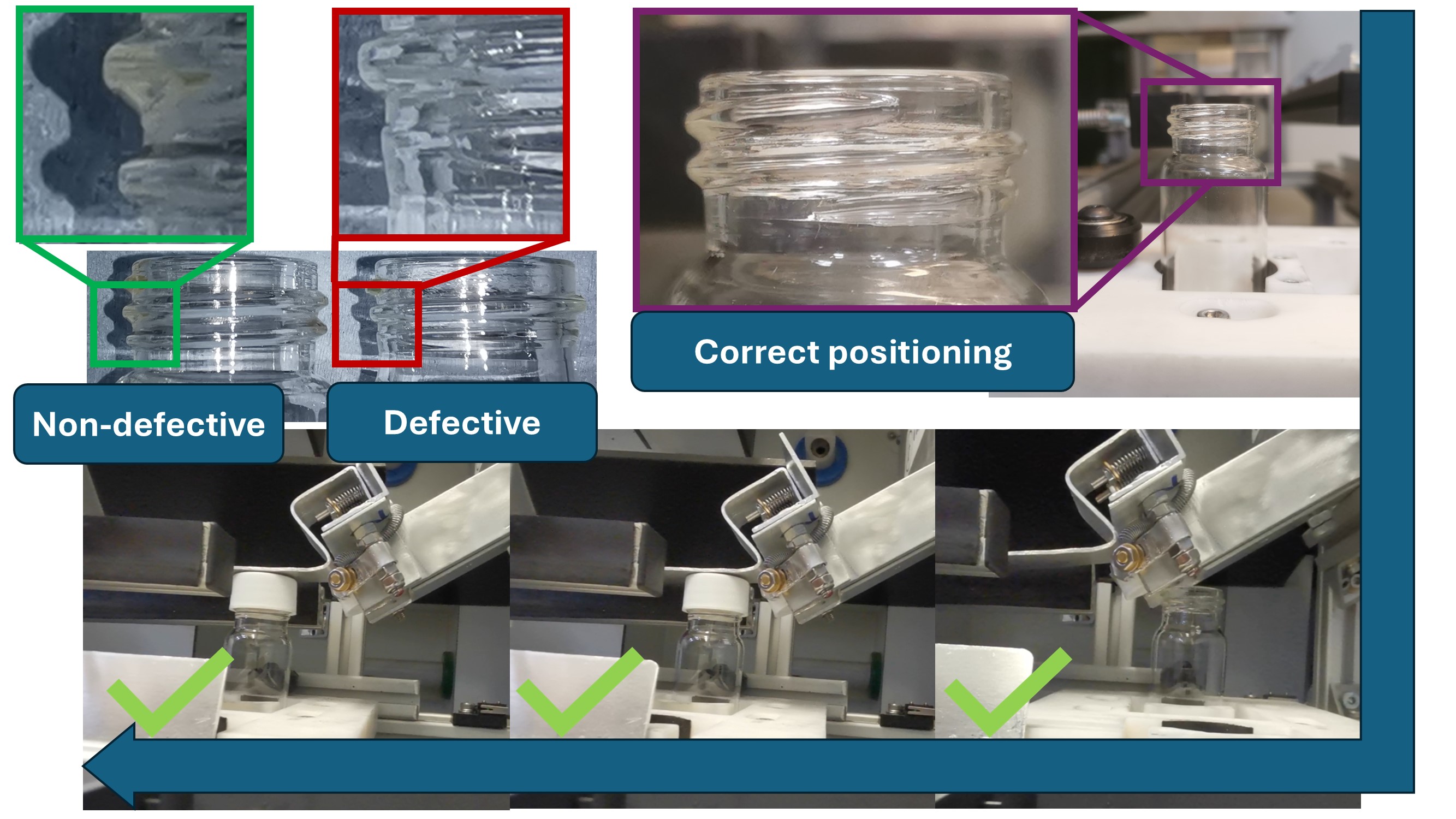}%
    \caption{Correct vial positioning and defects. A common defect can be observed in the top left of the image, where the vial on the left side is missing a thread. This defect leads to resistance or poor placement during capping and may cause the capping machine to fail to secure the cap. The correct positioning of the vial is shown in the top right corner; positioning the vial in this manner ensures that the threads of both the vial and cap engage properly. The bottom image shows the correct placement of the cap from the hopper onto a vial.}
    \label{fig:positioning}
\end{figure}

The experiments aim to measure how reliable the proposed machine is compared to manual or industrial alternatives. Therefore, two experiments are carried out: the first measures the prototype's capping and uncapping success rate, and the second measures the sealing capacities of the three approaches. The experimental setup is illustrated in Fig.~\ref{fig:expsetup}, a Panda robot is utilised to manipulate the vial and move them between different experimental sections. These stations are a balance (Quantos \textit{Mettler Toledo}, a pump for liquid dispensing (XCalibur \textit{Tecan Cavro} and the capping machine placed. The whole setup is confined within a fume cupboard. 
For the first experiment, the vials are first pre-selected and manually positioned at the orientation illustrated in Fig.~\ref{fig:positioning}. The robot then places the vials in the capping lane, and if the capping is successful, it moves the vial to the uncapping lane. This process is repeated 100 times (see Algorithm 2).

\begin{algorithm}[t]
\scriptsize
\caption{Vial Capping and Uncapping Process}
\begin{algorithmic}[1]
\For{$i = 1$ to $100$}
    \State Pick vial from rack and place in holder
    \State Machine caps the vial
    \State Check capping status
    \If {capping successful}
        \State Move vial to uncapping lane
    \Else
        \State Return cap, user uncaps manually
        \State \textbf{continue} (skip to next iteration)
    \EndIf
    \State Machine uncaps vial
    \State Remove cap with robot
    \State Return vial to rack
    \If {cap still on vial}
        \State Remove manually
    \EndIf
\EndFor
\end{algorithmic}
\end{algorithm}

In the second experiment, three batches of vials are prepared using three solvents—distilled water, ethanol, and acetone—with four repeats each. These solvents were chosen for their common use in chemistry and their varying vapour pressures, from low (water) to high (acetone)~\cite{waterPubChem,acetonePubChem,ethanolPubChem}. Since high-vapour-pressure solvents evaporate more readily, they present a greater challenge for effective sealing. Each vial is filled with 10 mL of solvent before capping. The first batch is capped manually, the second is prepared and capped by the Chemspeed system, and the third is prepared and capped by the robot. All batches are monitored over three days, with vial weights logged every three hours, to compare liquid loss across the three methods.

\section{Results}
\label{sec:results}

\begin{figure}[t!]
    \centering
    \includegraphics[width=3.3in]{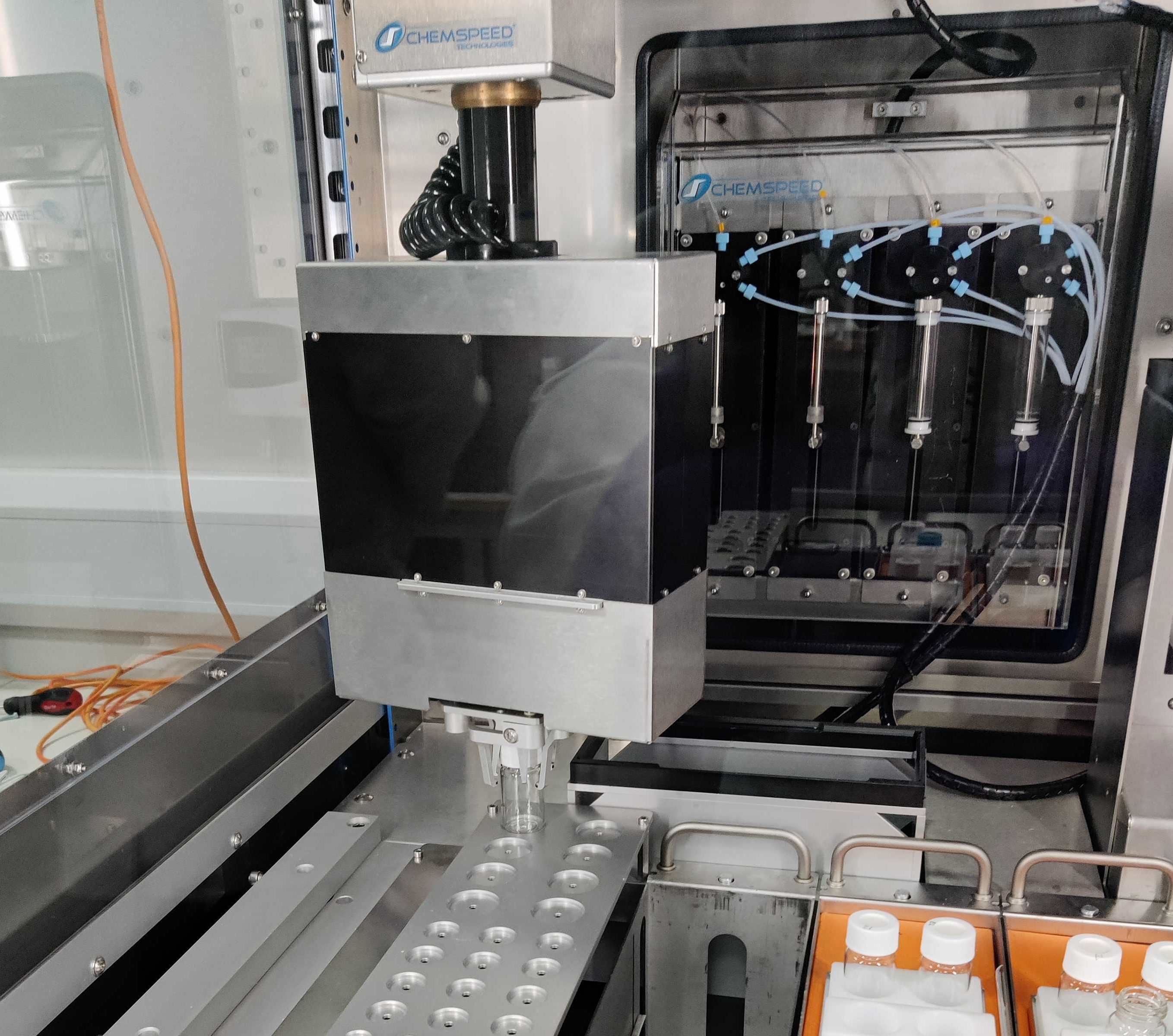}%
    \caption{The Chemspeed station failing to cap a defective vial. Normally, the caps rack is where the manipulator attempts to place the caps. However, due to an unsuccessful uncapping attempt, the manipulator attempted to place the entire vial there, resulting in a crash.}
    \label{fig:cheems}
\end{figure}

The results from both experiments revealed several insights into how to enhance the prototype's efficiency. In the first experiment, it was observed that the orientation of the vials in the rack significantly impacted the success rate. This is because of the importance of the orientation of the vial threads relative to the capping machine. Specifically, when the three threads were oriented toward the cap feeder (see Fig.~\ref{fig:positioning}), the success rate was 100 \%. This performance can be attributed to the increased likelihood that the internal threads of the cap engage with the vial threads, thereby facilitating the proper placement of the cap. This correct alignment enables the mechanism to spin and effectively tighten the cap. Moreover, it was found that as long as the prototype successfully caps the vial, uncapping also achieves a 100 \% success rate. During testing it was observed that a minority of vials were inconsistent or defective in the dimensions and number of screw threads on their necks. A common defect observed was a reduced number of vial threads. The capping machine was found to be unable to reliably cap these defective vials Fig.~\ref{fig:cheems}. However, simple visual analysis and pre-selection of non-defective vials was found to lead to reliable capping.

\begin{table*}[t!]
\scriptsize
\centering
\renewcommand{\arraystretch}{1.2} 
\setlength{\tabcolsep}{8pt} 
\begin{adjustbox}{max width=\textwidth}
\begin{tabular}{|l|c|c|c|c|c|c|c|c|c|}
\hline
\multirow{2}{*}{\makecell{Solvent \\ (sample)}} & \multirow{2}{*}{\makecell{Initial \\ weight (g)}} & \multirow{2}{*}{\makecell{Final \\ weight (g)}} & \multirow{2}{*}{\makecell{Density \\ (kg/m³)}} & \multirow{2}{*}{\makecell{Liquid \\ weight (g)}} & \multirow{2}{*}{\makecell{Weight \\ loss (mg)}} & \multirow{2}{*}{\makecell{Weight \\ loss (\%)}} & \multicolumn{3}{c|}{\textbf{Averages}} \\ \cline{8-10}
                     &                      &                      &                     &                     &                      &                       & \makecell{3 days \\ (\%)} & \makecell{Per day \\ (\%)} & \makecell{Per hour \\ (\%)} \\ \hline
Water (1)   & 29.2223  & 29.222   & \multirow{4}{*}{997}  & \multirow{4}{*}{9.97}  & 0.3   & 0.0030  & \multirow{4}{*}{0.0967}  & \multirow{4}{*}{0.0322}  & \multirow{4}{*}{0.0013}  \\ \cline{1-3} \cline{6-7}
Water (2)   & 28.158   & 28.1457  &      &       & 12.3  & 0.1233   &              &              &              \\ \cline{1-3} \cline{6-7}
Water (3)   & 28.0036  & 27.9843  &      &       & 19.3  & 0.1935  &              &              &              \\ \cline{1-3} \cline{6-7}
Water (4)   & 29.0125  & 29.0058  &      &       & 6.7   & 0.0672  &              &              &              \\ \hline
Ethanol (1) & 27.0353  & 27.0338  & \multirow{4}{*}{789}  & \multirow{4}{*}{7.89}  & 1.5   & 0.0190  & \multirow{4}{*}{0.6108}  & \multirow{4}{*}{0.2036}  & \multirow{4}{*}{0.0084}   \\ \cline{1-3} \cline{6-7}
Ethanol (2) & 27.3768  & 27.3748  &      &       & 2     & 0.0253  &              &              &              \\ \cline{1-3} \cline{6-7}
Ethanol (3) & 27.0333  & 26.8924  &      &       & 140.9 & 1.7858  &              &              &              \\ \cline{1-3} \cline{6-7}
Ethanol (4) & 26.27    & 26.2216  &      &       & 48.4  & 0.6134  &              &              &              \\ \hline
Acetone (1) & 26.9868  & 26.9542  & \multirow{4}{*}{784}  & \multirow{4}{*}{7.84}  & 32.6  & 0.4158  & \multirow{4}{*}{4.1868}  & \multirow{4}{*}{1.3956}  & \multirow{4}{*}{0.0581}  \\ \cline{1-3} \cline{6-7}
Acetone (2) & 24.871   & 24.7033  &      &       & 167.7 & 2.1390  &              &              &              \\ \cline{1-3} \cline{6-7}
Acetone (3) & 24.9481  & 23.8458  &      &       & 1102.3& 14.0599 &              &              &              \\ \cline{1-3} \cline{6-7}
Acetone (4) & 24.6049  & 24.5945  &      &       & 10.4  & 0.1326  &              &              &              \\ \hline
\end{tabular}
\end{adjustbox}
\caption{Capping Machine Data.}
\label{table:cappingmachine}
\end{table*}

\begin{table*}[t]
\scriptsize
\centering
\renewcommand{\arraystretch}{1.2} 
\setlength{\tabcolsep}{8pt} 
\begin{adjustbox}{max width=\textwidth}
\begin{tabular}{|l|c|c|c|c|c|c|c|c|c|}
\hline
\multirow{2}{*}{\makecell{Solvent \\ (sample)}} & \multirow{2}{*}{\makecell{Initial \\ weight (g)}} & \multirow{2}{*}{\makecell{Final \\ weight (g)}} & \multirow{2}{*}{\makecell{Density \\ (kg/m³)}} & \multirow{2}{*}{\makecell{Liquid \\ weight (g)}} & \multirow{2}{*}{\makecell{Weight \\ loss (mg)}} & \multirow{2}{*}{\makecell{Weight \\ loss (\%)}} & \multicolumn{3}{c|}{\textbf{Averages}} \\ \cline{8-10}
                     &                      &                      &                     &                     &                      &                       & \makecell{3 days \\ (\%)} & \makecell{Per day \\ (\%)} & \makecell{Per hour \\ (\%)} \\ \hline
Water (1)   & 27.972   & 27.9702  & \multirow{4}{*}{997}  & \multirow{4}{*}{9.97}  & 1.8   & 0.0181  & \multirow{4}{*}{0.0186}  & \multirow{4}{*}{0.0062}  & \multirow{4}{*}{0.0003}  \\ \cline{1-3} \cline{6-7}
Water (2)   & 29.2299  & 29.2279  &      &       & 2     & 0.0201  &              &              &              \\ \cline{1-3} \cline{6-7}
Water (3)   & 28.9663  & 28.9646  &      &       & 1.7   & 0.0171  &              &              &              \\ \cline{1-3} \cline{6-7}
Water (4)   & 28.1985  & 28.1966  &      &       & 1.9   & 0.0191  &              &              &              \\ \hline
Ethanol (1) & 26.0616  & 26.0601  & \multirow{4}{*}{789}  & \multirow{4}{*}{7.89}  & 1.5   & 0.0190  & \multirow{4}{*}{0.0212}  & \multirow{4}{*}{0.0071}  & \multirow{4}{*}{0.0003}   \\ \cline{1-3} \cline{6-7}
Ethanol (2) & 25.9324  & 25.9306  &      &       & 1.8   & 0.0228  &              &              &              \\ \cline{1-3} \cline{6-7}
Ethanol (3) & 24.7426  & 24.7411  &      &       & 1.5   & 0.0190  &              &              &              \\ \cline{1-3} \cline{6-7}
Ethanol (4) & 26.1948  & 26.1929  &      &       & 1.9   & 0.0241  &              &              &              \\ \hline
Acetone (1) & 26.0808  & 26.0778  & \multirow{4}{*}{784}  & \multirow{4}{*}{7.84}  & 3     & 0.0383  & \multirow{4}{*}{0.0456}  & \multirow{4}{*}{0.0152}  & \multirow{4}{*}{0.0006}  \\ \cline{1-3} \cline{6-7}
Acetone (2) & 25.002   & 24.9988  &      &       & 3.2   & 0.0408  &              &              &              \\ \cline{1-3} \cline{6-7}
Acetone (3) & 26.0644  & 26.0596  &      &       & 4.8   & 0.0612  &              &              &              \\ \cline{1-3} \cline{6-7}
Acetone (4) & 25.917   & 25.9137  &      &       & 3.3   & 0.0421  &              &              &              \\ \hline
\end{tabular}
\end{adjustbox}
\caption{Manual Capping Data.}
\label{table:manual}
\end{table*}

The results of the second experiment are summarised in Table~\ref{table:cappingmachine}, Table~\ref{table:manual} and Table~\ref{table:cheems} for the capping machine, manual capping and the Chemspeed, respectively. The vials containing water capped by the capping machine lost 0.0322 \% of weight on average per day, while those containing ethanol and acetone lost 0.2036 \%  and 1.39 \%, respectively.  The vials containing water capped manually lost 0.006\% of weight on average per day, while those containing ethanol and acetone lost 0.007 \% and 0.0152 \%, respectively. Lastly, the vials capped by the Chemspeed containing water lost 0.0123 \% of weight on average per day, while those containing ethanol and acetone lost 0.0053 \% and 0.014 \%, respectively.

\begin{table*}[t]
\scriptsize
\centering
\renewcommand{\arraystretch}{1.2} 
\setlength{\tabcolsep}{8pt} 
\begin{adjustbox}{max width=\textwidth}
\begin{tabular}{|l|c|c|c|c|c|c|c|c|c|}
\hline
\multirow{2}{*}{\makecell{Solvent \\ (sample)}} & \multirow{2}{*}{\makecell{Initial \\ weight (g)}} & \multirow{2}{*}{\makecell{Final \\ weight (g)}} & \multirow{2}{*}{\makecell{Density \\ (kg/m³)}} & \multirow{2}{*}{\makecell{Liquid \\ weight (g)}} & \multirow{2}{*}{\makecell{Weight \\ loss (mg)}} & \multirow{2}{*}{\makecell{Weight \\ loss (\%)}} & \multicolumn{3}{c|}{\textbf{Averages}} \\ \cline{8-10}
                     &                      &                      &                     &                     &                      &                       & \makecell{3 days \\ (\%)} & \makecell{Per day \\ (\%)} & \makecell{Per hour \\ (\%)} \\ \hline
Water (1)   & 28.3205  & 28.3199  & \multirow{4}{*}{997}  & \multirow{4}{*}{9.97}  & 0.6   & 0.00618  & \multirow{4}{*}{0.0123}  & \multirow{4}{*}{0.0041}  & \multirow{4}{*}{0.00017}  \\ \cline{1-3} \cline{6-7}
Water (2)   & 28.3858  & 28.3854  &      &       & 0.4      & 0.00401       &              &              &              \\ \cline{1-3} \cline{6-7}
Water (3)   & 28.2491  & 28.2482  &      &       & 0.9      & 0.00902       &              &              &              \\ \cline{1-3} \cline{6-7}
Water (4)   & 28.4515  & 28.4485  &      &       & 3      & 0.03001       &              &              &              \\ \hline
Ethanol (1) & 26.2059   & 26.2039   & \multirow{4}{*}{789}  & \multirow{4}{*}{7.89}  & 2      & 0.0253       & \multirow{4}{*}{0.0161}       & \multirow{4}{*}{0.0053}       & \multirow{4}{*}{0.00022}       \\ \cline{1-3} \cline{6-7}
Ethanol (2) & 25.3897  &  25.3884  &      &       & 1.3      & 0.0164       &              &              &              \\ \cline{1-3} \cline{6-7}
Ethanol (3) & 26.3091   &  26.3080  &      &       & 1.1      & 0.0139       &              &              &              \\ \cline{1-3} \cline{6-7}
Ethanol (4) & 25.1165  & 25.1158  &      &       & 0.7      & 0.0088       &              &              &              \\ \hline
Acetone (1) & 25.9171  & 25.914   & \multirow{4}{*}{784}  & \multirow{4}{*}{7.84}  & 3.1    & 0.0395  & \multirow{4}{*}{0.0424}  & \multirow{4}{*}{0.0141}  & \multirow{4}{*}{0.0006}  \\ \cline{1-3} \cline{6-7}
Acetone (2) & 25.8261  & 25.8227  &      &       & 3.4    & 0.0434  &              &              &              \\ \cline{1-3} \cline{6-7}
Acetone (3) & 26.1036  & 26.1001  &      &       & 3.5    & 0.0446  &              &              &              \\ \cline{1-3} \cline{6-7}
Acetone (4) & 25.8148  & 25.8115  &      &       & 3.3    & 0.0421  &              &              &              \\ \hline
\end{tabular}
\end{adjustbox}
\caption{Chemspeed Data.}
\label{table:cheems}
\end{table*}

For the three approaches, the average weight loss per day, considering the three solvents, was 0.54 \% for the capping machine, 0.013 \% for manual capping and 0.0078 \% for the Cheemspeed. These results show that acetone is the most challenging solvent to contain due to its lower boiling point and higher volatility than water and ethanol. Moreover, the torque control that the Chemspeed implements secured better sealing capacities for the three solvents. However, the Chemspeed station cannot detect capping failure Fig.~\ref{fig:cheems}. Manual capping proved more consistent in this context. Notorious differences in tightness of fit can be detected by humans during capping of different vials, however,  they are easily compensated for by attentive users by adjusting torque or changing caps.

\section{Discussion}
\label{sec:discussion}

The results showed that, despite industry implementing capping for decades, even industrial cappers are not free from errors, which are often provoked by unavoidable variations in the quality of the vials and caps. Much has been said about robots and autonomous systems being less susceptible to making errors, however, not much is usually said about how humans can overcome the mistakes these systems often provoke. In contrast, automated systems are incapable of, if not designed or programmed to handle unexpected events. In the context of the experiments carried out herein, manual capping was more reliable in identifying failing capping and defective parts, moreover humans are more adaptable and can adjust the needed force to cap in each individual case; therefore,  manual sealing was more consistent than its automated counterparts.

Although Chemspeed uses a closed-loop torque system to detect capping issues, defective vials or caps can cause torque misreadings due to slight manufacturing variations. To address this, our capping machine integrates a vision-based decision framework for added robustness. It offers a compact, dedicated alternative to commercial machines, avoiding unnecessary robotic wear and making the process robot-agnostic. Unlike workflows where robots directly cap vials~\cite{lunt2024pxrd}, our modular design fits various robot types (\textit{e.g.}, cartesian, SCARA, arm) with minimal adaptation, as long as they can place and retrieve vials. While currently limited to 20 ml vials, expanding to multiple lanes and feeders could enhance flexibility and support various sizes, all while maintaining a compact footprint.

\section{Conclusions and Future Work}
\label{sec:conclusion}

This paper has presented a prototype of a capping machine suitable for confined spaces. The machine proved reliable by achieving a 100 \%  success rate in capping/uncapping and losing around 0.54 \%  of weight after one day of monitoring the vials' mass---acetone was the most challenging to contain among all solvents, even for the industrial capper. This loss is not significant for most chemical experiments in SDLs, as it is primarily due to vapour evaporation rather than leakage. Therefore, as long as the vial is not leaking, processes such as mixing and temporary sample storage can be carried out safely. Future work can potentially focus on increasing the flexibility of the capping machine. This would require adding more lanes and cap hoppers to support a broader range of vials and caps. The vision module can also be further improved to recognise the type of vial put in the capper, and whether it is defective and properly oriented. Additionally, adding an extra actuator that tightens the caps with torque feedback would increase the capping machine's robustness. Finally, exploring approaches to solve or react to unexpected or novel situations, such as broken vials and dangerous liquid spills or leaks, is fundamental for securing safe and robust automation in SDLs.

\begin{credits}
\subsubsection{\ackname} This project was funded by the ERC ADAM Synergy grant (agreement no. 856405), the Engineering and Physical Sciences Research Council (EPSRC) under the grant agreement EP/V026887/1 and the Leverhulme Trust through the Leverhulme Research Centre for Functional Materials Design.

\subsubsection{\discintname}
The authors declare no competing interests. 
\end{credits}
%
%
%
%
\bibliographystyle{splncs04}
\bibliography{rsc}

\begin{thebibliography}{10}
\providecommand{\url}[1]{\texttt{#1}}
\providecommand{\urlprefix}{URL }
\providecommand{\doi}[1]{https://doi.org/#1}

\bibitem{Li2015SmallMolecule}
Li, J., Ballmer, S.G., Gillis, E.P., Fujii, S., Schmidt, M.J., Palazzolo, A.M.E., Lehmann, J.W., Morehouse, G.F., Burke, M.D.: Synthesis of many different types of organic small molecules using one automated process. Science  \textbf{347}(6227),  1221--1226 (2015). \doi{10.1126/science.aaa5414}, \url{https://www.science.org/doi/abs/10.1126/science.aaa5414}

\bibitem{Salley2023Chemput}
Salley, D., Manzano, J.S., Kitson, P.J., Cronin, L.: Robotic modules for the programmable chemputation of molecules and materials. ACS Central Science  \textbf{9}(8),  1525--1537 (2023). \doi{10.1021/acscentsci.3c00304}, \url{https://doi.org/10.1021/acscentsci.3c00304}

\bibitem{burger2020mobile}
Burger, B., Maffettone, P.M., Gusev, V.V., Aitchison, C.M., Bai, Y., Wang, X., Li, X., Alston, B.M., Li, B., Clowes, R., et~al.: A mobile robotic chemist. Nature  \textbf{583}(7815),  237--241 (2020)

\bibitem{Wu2023Laser}
Wu, T.C., Aguilar-Granda, A., Hotta, K., Yazdani, S.A., Pollice, R., Vestfrid, J., Hao, H., Lavigne, C., Seifrid, M., Angello, N., Bencheikh, F., Hein, J.E., Burke, M., Adachi, C., Aspuru-Guzik, A.: A materials acceleration platform for organic laser discovery. Advanced Materials  \textbf{35}(6),  2207070 (2023). \doi{https://doi.org/10.1002/adma.202207070}, \url{https://onlinelibrary.wiley.com/doi/abs/10.1002/adma.202207070}

\bibitem{10093437}
Paul, J.E., G, L.: Cost efficient automatic filling system for differently sized bottles. In: 2022 International Conference on Recent Trends in Microelectronics, Automation, Computing and Communications Systems (ICMACC). pp.~1--6 (2022). \doi{10.1109/ICMACC54824.2022.10093437}

\bibitem{kumar2014automation}
Kumar, A.A., Rao, P.S.: Automation of bottle manufacturing, filling and capping process using low cost industrial automation. International Journal of Engineering Research \& Technology (IJERT), ISSN pp. 2278--0181 (2014)

\bibitem{yao2023design}
Yao, Q., Chai, C., Liu, X., Pang, W., Shen, D., Cao, X.: Design of pneumatic control system for automatic bottle capping device. In: Third International Conference on Mechanical Design and Simulation (MDS 2023). vol. 12639, pp. 35--43. SPIE (2023)

\bibitem{mahrez2022design}
Mahrez, A., Hilmy, A., Alasoad, A., Yahya, A.E., Yahya, K., Amer, A., Attar, H.: Design a plc-based automated and controlled liquid filling-capping system. In: 2022 International Engineering Conference on Electrical, Energy, and Artificial Intelligence (EICEEAI). pp.~1--5. IEEE (2022)

\bibitem{zhang2020control}
Zhang, S., Ji, J., Zhao, Y., Li, Y.: Control system design of automatic capping machine based on s7-300 plc. In: 2020 IEEE Conference on Telecommunications, Optics and Computer Science (TOCS). pp. 337--339. IEEE (2020)

\bibitem{jaeger2021automated}
Jaeger, J.W., Adkins, S.C., Perez-Tamayo, S.C., Werth, K.E., Hansen, G., Nimunkar, A.J., Radwin, R.G.: Automated device for uncapping multiple-size bioanalytical sample tubes designed to reduce technician strain and increase productivity. SLAS TECHNOLOGY: Translating Life Sciences Innovation  \textbf{26}(3),  320--326 (2021)

\bibitem{kumar2016design}
Kumar, A., Pillearachichige, K., Sharifi, H., Shaw, B., Noble, F.K.: Design of end-effectors for a chemistry automation plant. In: 2016 23rd International Conference on Mechatronics and Machine Vision in Practice (M2VIP). pp.~1--5. IEEE (2016)

\bibitem{ZoneSunCappers}
LIMITED, Z.T.: Zonesun capping machines. \url{https://www.zonesun.com/collections/cap-screwing}, accessed: 25-11-2024

\bibitem{AzentaIntellicap}
SCIENCES, A.L.: Azenta intellixcap™. \url{https://www.azenta.com/intellixcap-tube-capping-decapping-sealing-systems?}, accessed: 25-11-2024

\bibitem{waterPubChem}
for Biotechnology Information~(2024), N.C.: Pubchem compound summary for cid 962, water. https://pubchem.ncbi.nlm.nih.gov/compound/Water, retrieved September 20, 2024

\bibitem{acetonePubChem}
for Biotechnology Information~(2024), N.C.: Pubchem compound summary for cid 180, acetone. https://pubchem.ncbi.nlm.nih.gov/compound/Acetone, retrieved September 20, 2024

\bibitem{ethanolPubChem}
for Biotechnology Information~(2024), N.C.: Pubchem compound summary for cid 702, ethanol. https://pubchem.ncbi.nlm.nih.gov/compound/Ethanol, retrieved September 20, 2024

\bibitem{lunt2024pxrd}
Lunt, A.M., Fakhruldeen, H., Pizzuto, G., Longley, L., White, A., Rankin, N., Clowes, R., Alston, B., Gigli, L., Day, G.M., Cooper, A.I., Chong, S.Y.: Modular, multi-robot integration of laboratories: an autonomous workflow for solid-state chemistry". Chem. Sci.  \textbf{15},  2456--2463 (2024). \doi{10.1039/D3SC06206F}, \url{http://dx.doi.org/10.1039/D3SC06206F}

\end{thebibliography}
\end{document}